\documentclass[runningheads]{llncs}

\usepackage[T1]{fontenc}
\usepackage{cite}
\usepackage{graphicx}
\usepackage{booktabs}
\usepackage{url}
\usepackage{amsfonts}
\usepackage{nicefrac}
\usepackage{microtype}
\usepackage{float}
\usepackage{amsmath}
\usepackage{pifont}
\usepackage{multirow}
\usepackage{tabularx}
\usepackage{caption}
\usepackage{makecell}

\usepackage{xcolor}         % colors

\begin{document}

\title{Mil-SCORE: Benchmarking Long-Context Geospatial Reasoning and Planning in Large Language Models}

\titlerunning{MilSCORE}

\author{Aadi Palnitkar\inst{1} \and
Mingyang Mao\inst{2,4} \and
Nicholas Waytowich\inst{3} \and Vinicius G. Goecks\inst{3} \and Xiaomin Lin\inst{2,4}}

\authorrunning{A. Palnitkar et al.}

\institute{University of Maryland, College Park MD, USA \and
ERA Lab, University of South Florida, Tampa FL, USA \and
DEVCOM Army Research Laboratory, Aberdeen Proving Ground MD, USA \and
EEHPC Lab, Johns Hopkins University, Baltimore MD, USA}

\maketitle

\begin{abstract}
As large language models (LLMs) are applied to increasingly longer and more complex tasks, there is a growing need for realistic long-context benchmarks that require selective reading and integration of heterogeneous, multi-modal information sources. This need is especially acute for geospatial planning problems, such as those found in planning for large-scale military operations, which demand fast and accurate reasoning over maps, orders, intelligence reports, and other distributed data. To address this gap, we present MilSCORE (Military Scenario Contextual Reasoning), to our knowledge the first scenario-level dataset of expert-authored, multi-hop questions grounded in a complex, simulated military planning scenario used for training. MilSCORE is designed to evaluate high-stakes decision-making and planning, probing LLMs' ability to combine tactical and spatial reasoning across multiple sources and to reason over long-horizon, geospatially rich context. The benchmark includes a diverse set of question types across seven categories targeting both factual recall and multi-step reasoning about constraints, strategy, and spatial analysis. We provide an evaluation protocol and report baseline results for a range of contemporary vision-language models. Our findings highlight substantial headroom on MilSCORE, indicating that current systems struggle with realistic, scenario-level long-context planning, and positioning MilSCORE as a challenging testbed for future work.

\end{abstract}

\begin{figure}[h]
\includegraphics[width=\textwidth]{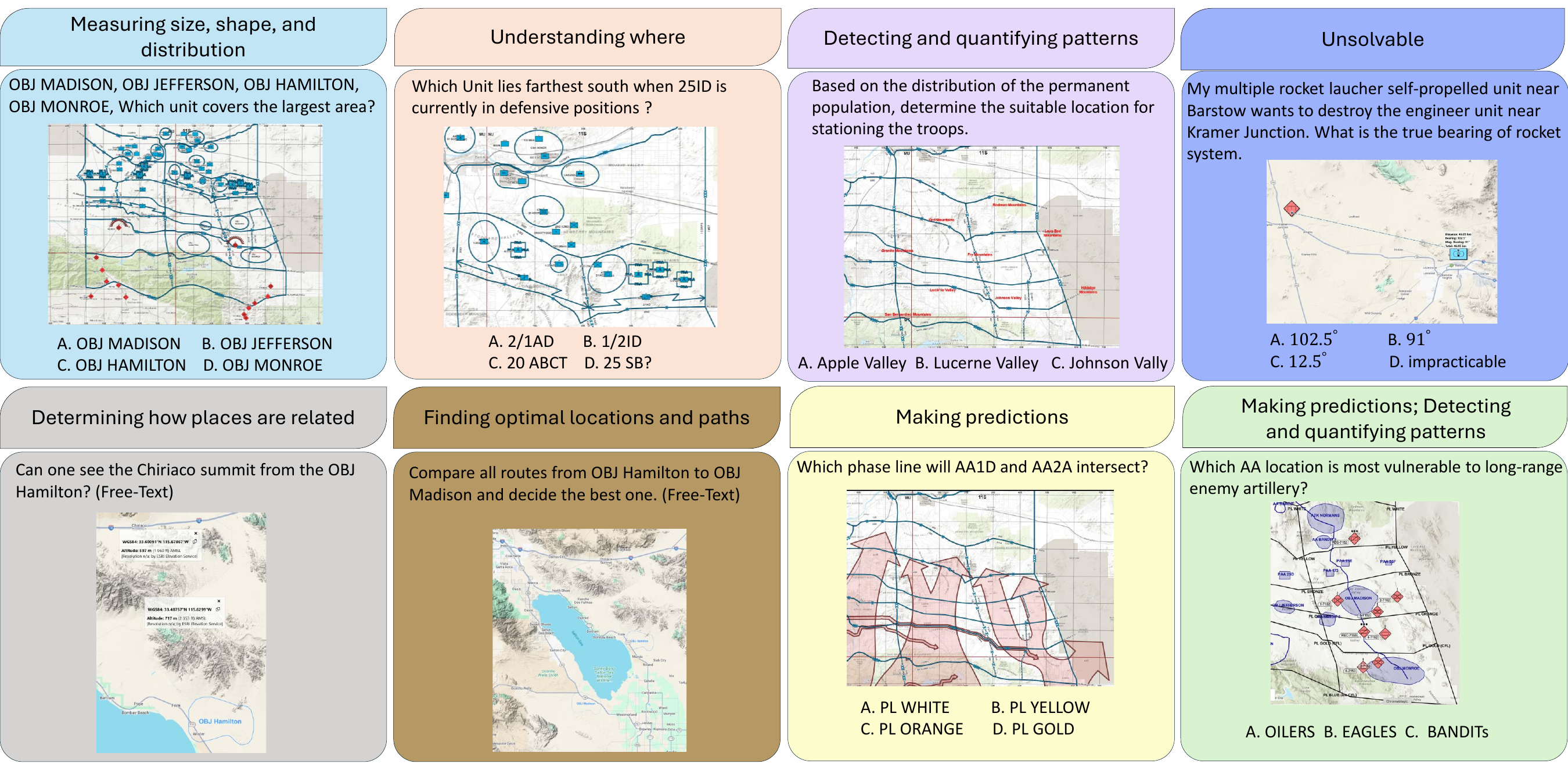}
\caption{Example MilSCORE questions grouped by spatial-analysis content categories. The figure illustrates representative items from each category: Understanding where; Measuring size, shape, and distribution; Determining how places are related; Finding optimal locations and paths; Detecting and quantifying patterns; Making predictions; and Unsolvable Tasks---including both multiple-choice and free-text formats, reflecting the diversity and complexity of real-world military geospatial reasoning scenarios.}
\label{fig2.questions}
\end{figure}

\section{Introduction}

Geospatial information analytics are widely used for multi-domain downstream tasks, including transportation\cite{GISforTransportation}, agriculture\cite{GISforAgriculture}, public health\cite{FradelosHealthBasedGeographic,Wang02012020}, and also in high-stakes scenarios such as natural disaster response and military operations\cite{MAI2025104368,ijgi1020166,fleming2009role,goecks2023disasterresponsegptlargelanguagemodels}. Innovations in remote sensing, satellite systems, LiDAR, and GPS technologies have made geospatial data generation more affordable, while advances in machine learning and computer vision now enable more efficient extraction of valuable information\cite{Selmy24}. However, the subsequent geometric planning and reasoning process integrates geometric coordinates, maps, satellite imagery, and socioeconomic data, a task that has traditionally depended on substantial human expertise and manual effort.

Advanced large language models (LLMs) and their multimodal extensions, vision-language models (VLMs), offer a new avenue for automating complex geospatial reasoning tasks and have demonstrated extensive world knowledge\cite{manvi2024geollmextractinggeospatialknowledge}. However, directly applying general-purpose LLMs to specialized domains such as military geospatial planning poses several challenges. One critical issue is hallucination: LLMs can produce incorrect or inconsistent geospatial information that is not grounded in reality\cite{wang2025mitigatinggeospatialknowledgehallucination}. Another challenge is the long-context nature of planning problems—operation orders and supporting documents are often very lengthy and encompass multiple heterogeneous sources. Current LLMs struggle with such extensive inputs, effectively utilizing only a fraction of very large context windows and often exhibiting performance degradation as context length grows\cite{kuratov2024babilongtestinglimitsllms}. These limitations highlight the need for realistic long-context benchmarks that reflect the demands of geospatial planning and reasoning in military operation order (OPORD) scenarios.

In this work, we study military decision-making tasks for VLMs using course-of-action (COA) maps paired with standardized geospatial feature data (GeoJSON) and associated OPORD documentation. We introduce \textbf{MilSCORE} (Military Scenario Contextual Reasoning), to our knowledge the first scenario-level dataset of expert-authored, multi-hop questions grounded in a complex training OPORD. MilSCORE encompasses over 100 comprehensive queries spanning over 50 distinct maps derived from a a complete operational scenario used for military training. The benchmark is designed to probe high-stakes decision-making and planning, including force allocation, synchronization across phases, and risk assessment under doctrinal constraints. We evaluate multiple proprietary and open-source VLMs on MilSCORE, including Llama-3.2 11B-Vision-Instruct, Qwen-2-VL-7B, GPT-4o, and GPT-4o mini under both zero-shot prompting and chain-of-thought reasoning protocols, and analyze their failure modes in long-context, geospatially grounded reasoning.

In summary, the contributions of this paper are:
\begin{enumerate}
    \item We introduce \textbf{MilSCORE}, a novel scenario-level benchmark of expert-authored, multi-hop questions and answers grounded in training operation orders and COA maps, designed to test the military decision-making process (MDMP) in realistic, long-context geospatial planning settings.
    \item We propose an evaluation protocol and provide a comprehensive empirical study of state-of-the-art VLMs on MilSCORE, establishing baseline results and identifying key challenges for future work on long-context, decision-focused geospatial reasoning with LLMs and VLMs.
\end{enumerate}

\section{Related Work}

\subsection{Enhancing reasoning and planning in pretrained models}

Pre-trained large VLMs have shown strong capabilities across a wide range of domains~\cite{wu2025dream,Elgendy2025ChatENVAI,marafioti2025smolvlm,LIU2025100943,brown2025alphaearthfoundationsembeddingfield}. Their performance can often be further improved by combining them with complementary techniques. Retrieval-augmented generation (RAG) mitigates the limits of context and static parametric knowledge by integrating external knowledge bases at inference time~\cite{luo2024videoragvisuallyalignedretrievalaugmentedlong, mao2025multirag}. More recent work that explicitly intertwines RAG and reasoning lets large models dynamically interleave retrieval and multi-step reasoning, producing more accurate, context-aware, and goal-driven solutions~\cite{gao2025_synergizing_rag_reasoning_systematic,islam2024openragenhancedretrievalaugmentedreasoning,li2025agenticragdeepreasoning}. Chain-of-thought (CoT) prompting improves factual grounding and compositional generalization by encouraging models to produce intermediate reasoning steps~\cite{wei2023chainofthoughtpromptingelicitsreasoning,yao2023treethoughtsdeliberateproblem}. Our evaluation protocol adopts this perspective and uses a lightweight tool-using CoT agent as the test harness over MilSCORE.

\subsection{Geographic and Spatial AI}

From high-level strategic synthesis to fine-grained spatial interpretation, large (vision) language models are increasingly used in geospatial AI. Applications range from semantic understanding of natural-language queries and image-grounded reasoning to the integration of external geographic knowledge graphs, retrieval-augmented generation pipelines, and spatial-aware modules that support localized predictions and domain-specific planning across diverse tasks and environments~\cite{GEOBench-VLM,xu2025geor1,zhang2025geor1,lee2025multiagentgeospatialcopilotsremote,singh2024geollmenginerealisticenvironmentbuilding,liu2025geographyawarelargelanguagemodels}. Foundation models over satellite and aerial imagery, such as AlphaEarth Foundations~\cite{brown2025alphaearthfoundationsembeddingfield}, provide powerful representations that can be paired with LLMs for downstream decision-making.

Spatial RAG methods extend vanilla RAG by integrating structured spatial databases with large language models through hybrid retrieval---combining sparse spatial filters and dense semantic matching---to enable precise, context-aware reasoning over real-world geographic queries that involve complex spatial relationships, such as proximity, direction, and region-based constraints~\cite{yu2025_spatial_rag_spatial_retrieval_augmented,yu2025raggeoscienceexpectgaps,LIANG2025104712}.

\subsection{Benchmarks on Geographic and Geospatial Tasks}

Benchmarks designed to evaluate geographic artificial intelligence tasks have shown a distinct progression, advancing from foundational single-modality tasks to complex, agent-style geospatial workflows. Early work focused on single tasks such as visual recognition and geospatial semantics, including GeoImageNet~\cite{GeoImageNet}, GeoLLM~\cite{manvi2024geollmextractinggeospatialknowledge}, and the language-only GeoGLUE benchmark for geographic text understanding~\cite{li2023geogluegeographiclanguageunderstanding}. 

Subsequently, a series of evaluations was introduced to investigate multi-modal, comprehensive cognitive tasks. Benchmarks such as SpatialEval~\cite{wang2024spatialeval}, GEOBench-VLM~\cite{GEOBench-VLM}, and research by Ji et al.~\cite{Ji_2025} and Xu et al.~\cite{xu2025evaluatinglargelanguagemodels} appraise models' comprehension of spatial relationships and multi-step compositional reasoning. In the latest phase, the focus has decisively shifted toward agentic capabilities. Advanced benchmarks, exemplified by GeoAnalystBench~\cite{GeoAnalystBench}, the Cloud-Based Geospatial Benchmark (CBGB)~\cite{Cloud-Based-Geospatial-Benchmark}, and GeoBenchX~\cite{GeoBenchX}, now evaluate LLMs not for passive reasoning, but as active agents capable of generating code, invoking external tools, and executing entire, complex geospatial analysis workflows. Our work is also inspired by MapBench~\cite{MapBench} and MapEval~\cite{MapEval}, which study visually and textually demanding map-reading tasks. MilSCORE contributes to this line of work by targeting long-context, scenario-level military decision-making grounded in realistic OPORD materials.

\newcolumntype{Y}{>{\centering\arraybackslash}X}
\begin{table}[htb]
  \scriptsize
  \setlength{\tabcolsep}{2pt}
  \begin{tabularx}{\textwidth}{l *{6}{Y}}
    \toprule
    \multirow{2}{*}{Benchmarks} & \multirow{2}{*}{Modality} & Answer & Annotation & Task  & Interac. & Multi \\
    & & Type & Type  & Categories & Tools & Steps  \\
    \midrule
    GeoAnalystBench\cite{GeoAnalystBench}       & Text  & WFS, Co & M & 50 & \ding{51} & \ding{51} \\\addlinespace
    GEOBench-VLM\cite{GEOBench-VLM}
      & T + G
      & MCQ, BBox, Seg
      & A+M
      & \makecell{8 (T1)\\31 (T2)}
      & \ding{55}
      & \ding{51} \\\addlinespace
    GeoBenchX\cite{GeoBenchX}                   & Text & FFT, SoT & M & 4  & \ding{51} & \ding{51}\\\addlinespace
    GeoImageNet\cite{GeoImageNet}               & Hybrid & BBOX & M & 6 & \ding{55}  & \ding{55} \\\addlinespace
    MapEval \cite{MapEval}                      & Hybrid & MCQ & A + M &  6 & \ding{51} & \ding{51}\\\addlinespace
    CBGB \cite{Cloud-Based-Geospatial-Benchmark}& Text  & SNA & A + M  & E, I, D & \ding{51} & \ding{51}  \\\addlinespace
    GeoHaluBench \cite{mitigatinggeospatialknowledgehallucination}
          & Text + KG
          & FFT
          & A + M
          & \makecell{3 (T1)\\5 (T2)\\21 (T3)}
          & \ding{51}
          & \ding{51} \\\addlinespace
    MapBench \cite{MapBench}                    & Text & FFT & M & 9 & \ding{51}  & \ding{51} \\\addlinespace
    \textbf{MilSCORE (ours)}               & Hybrid & MCQ, FFT &  M & 7  & \ding{51}  & \ding{51} \\
    \bottomrule
  \end{tabularx}
  \vspace{10pt}
  \caption{Comparison of existing geographic and map benchmarks. Details: modalities (T = Text, G = Graphs, KG = Knowledge Graph); answer type (WFS = workflow steps, Co = code, MCQ = multiple-choice questions, BBox = bounding box, Seg = segmentation mask, FFT = free-form text, SoT = sequence of tools, SNA = single numerical answer); annotation type (M = manual, A = automatic); task categories (E/I/D = easy, intermediate, difficult).}
  \label{sample-table}
\end{table}

\begin{figure}[h]
\includegraphics[width=\textwidth]{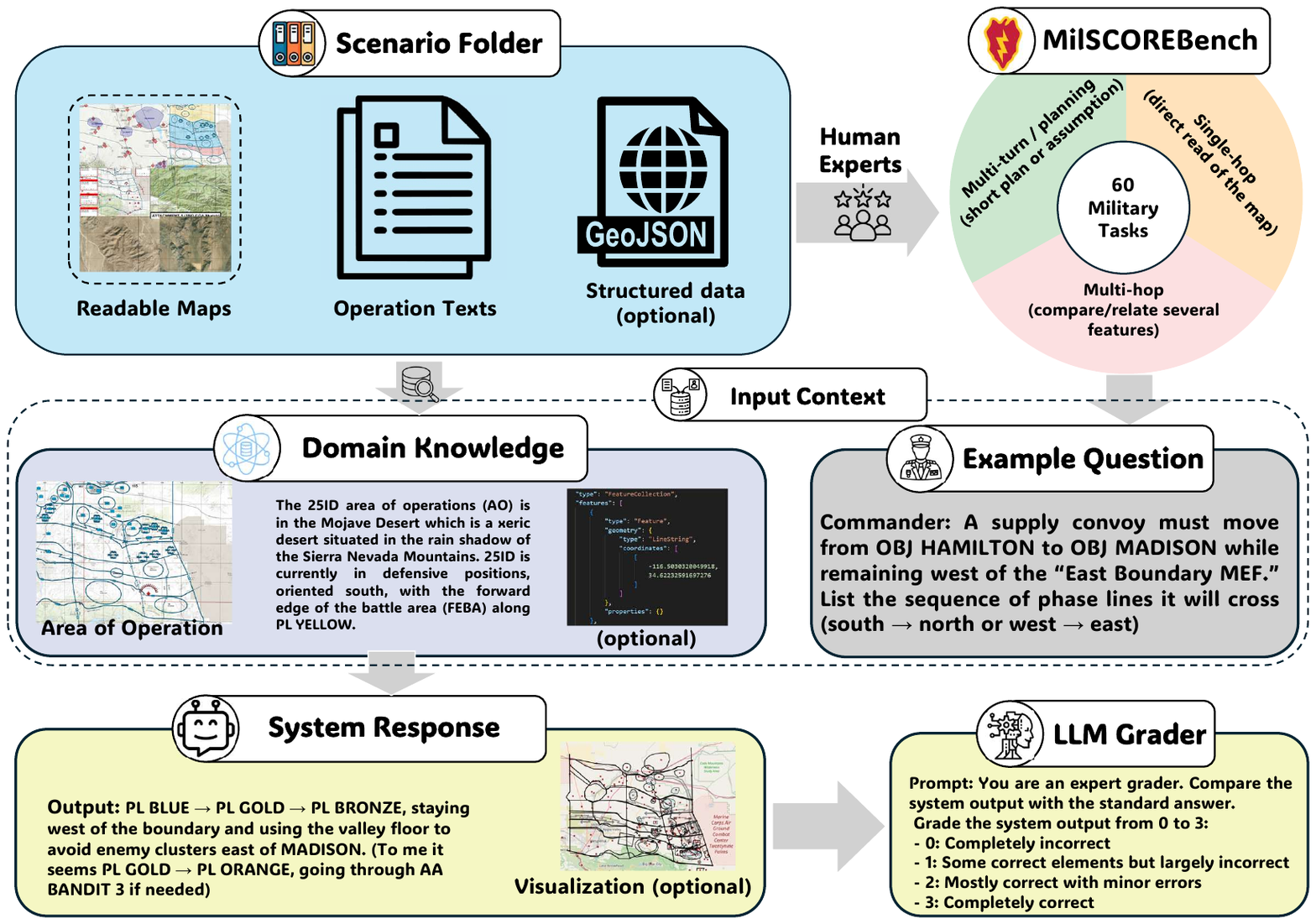}
\caption{Overview of the MilSCORE dataset and evaluation pipeline. Unclassified scenario folders containing readable maps, operation texts, and optional structured GeoJSON are used by human experts to author 50 military decision-making tasks spanning single-hop and multi-hop reasoning tiers. For evaluation, a tool-using chain-of-thought agent iteratively inspects the maps and documents, reasons about the question, and produces a free-text or structured answer (optionally with a visualization). An LLM-based grader then compares the final answer against the expert reference to assign a discrete correctness score.}
\label{fig1.benchoverview}
\end{figure}

\section{MilSCORE Dataset}

To develop a reliable, professional military geospatial benchmark, we first designed a comprehensive multi-modal dataset of realistic battlefield scenarios. The following sections detail the composition and taxonomy of this dataset, followed by our methodology for dataset construction, collection, and expert validation.

\subsection{Composition and Taxonomy}

According to their level of complexity, MilSCORE tasks are grouped into three difficulty tiers. Tier~1 comprises single-hop questions that can be answered directly by consulting a single information source, such as reading from a map or accessing domain-specific knowledge. Tier~2 consists of multi-hop questions that require comparison or relational reasoning over several features, while still relying on a single underlying information source. Tier~3 encompasses more complex multi-hop questions that necessitate the synthesis of a series of heterogeneous pieces of information---typically combining maps, GeoJSON overlays, and order text---with a particular emphasis on planning processes and hypothetical or assumption-based reasoning.

From the perspective of task content, and inspired by GeoAnalystBench~\cite{GeoAnalystBench}, we adopt ESRI's classic taxonomy of spatial analysis~\cite{esri2013language} to structure the categories in our benchmark. In addition, following MapEval~\cite{MapEval}, we introduce an \emph{Unsolvable Task} category. These tasks are intentionally constructed so that the available information is sufficient to conclude that the mission cannot be achieved, thereby allowing us to evaluate whether a model can correctly identify impossibility even when the prompt is intentionally misleading. Overall, our benchmark encompasses the following seven categories:
\begin{itemize}
    \item Understanding where,
    \item Measuring size, shape, and distribution,
    \item Determining how places are related,
    \item Finding optimal locations and paths,
    \item Detecting and quantifying patterns,
    \item Making predictions, and
    \item Unsolvable Tasks.
\end{itemize}
Each benchmark question may align with one or more of these categories, reflecting the multifaceted nature of real-world geospatial reasoning.

\subsection{Dataset Construction}

The MilSCORE dataset is constructed from unclassified military scenario folders. As shown in Figure~\ref{fig1.benchoverview}, these sources provide a realistic multi-modal environment, combining: (i) readable imagery including satellite remote sensing, terrain maps, and layered operation maps; (ii) textual documents such as operation orders and course-of-action briefings; and (iii) structured data on unit locations and boundaries (for example, in GeoJSON). This approach ensures the dataset captures the complexity of tactical planning tasks using only open-source information.

We also use Military Map~\cite{maparmy}, a modern, platform-independent web application for planning military exercises and missions, to manually draw the operational maps used in our benchmark, enabling us to construct standards-compliant military graphics that accurately reflect tactical planning workflows. Each question in our benchmark, along with its corresponding reference answer, is authored and then reviewed by experts with professional experience in military operations and geospatial analysis, ensuring the soundness of task design and the reliability of the provided solutions. Annotators also assign a difficulty tier and spatial-analysis category (or categories) to each item.

\subsection{Dataset Statistics}

The current MilSCORE release contains more than 100 expert-authored questions grounded in 50 distinct operational maps drawn from a single OPORD-style training scenario. Questions are distributed across the three difficulty tiers and seven spatial-analysis categories described above; Figure~\ref{fig2.questions} illustrates representative examples from each category. In our experiments (Section~\ref{sec:experiments}), we focus on a 60-question slice spanning all tiers to keep the study tractable while preserving the mix of map-reading and higher-level planning problems.

\section{Experiments}
\label{sec:experiments}

\begin{table}[htb]
  \scriptsize
  \setlength{\tabcolsep}{4pt}
  \begin{tabularx}{\textwidth}{l *{4}{Y}}
    \toprule
    \textbf{Tier} & \textbf{GPT-4o} & \textbf{Claude Sonnet 4.5} & \textbf{Claude Haiku 4.5} & \textbf{Gemini 2.5 Flash} \\
    \midrule
    Tier~1 (single-hop) (20)             & 8 (40\%)  & 6 (30\%) & 6 (30\%) & 8 (40\%) \\
    Tier~2 (single-source multi-hop) (20) & 12 (60\%) & 7 (35\%) & 4 (20\%)  & 9 (45\%) \\
    Tier~3 (cross-source multi-hop) (20) & 15 (75\%) & 0 (0.0\%)  & 6 (30\%) & 14 (70\%) \\
    \midrule
    \textbf{Total}                   & \textbf{35 (58.3\%)} & \textbf{13 (21.7\%)} & \textbf{16 (26.7\%)} & \textbf{31 (51.7\%)} \\
    \bottomrule
  \end{tabularx}
  \vspace{6pt}
  \caption{Results on a 60-question MilSCORE slice. Each entry reports the estimated number of correctly answered questions in a given difficulty tier (percentages are taken with respect to the number of questions tested in each tier).}
  \label{tab:complexity-acc}
\end{table}

GPT-4o leads the slice because it is the only model that consistently finishes multi-step loops within the 10-step cap we impose to keep API costs bounded. Larger “reasoning” variants such as Claude Sonnet 4.5 and Gemini 2.5 Flash tend to narrate at length, spending multiple turns describing intent before touching a tool; they therefore burn through the 10-step budget before the evidence collection is complete, yielding many “max iterations reached” errors. GPT-4o is terse, immediately chains tools, and thus converts more prompts into boxed answers. We initially attempted to include OpenAI’s GPT-5 preview, but the preview API throttles tool-heavy loops and rejects native image payloads, so we excluded it from the reported table.

Interestingly, Tier~3 accuracies beat Tier~1/Tier~2 for the models that survive the loop. Cross-source prompts practically force a “list → read PDF → query spreadsheet → inspect map” pattern, which aligns with our tool instructions and keeps the model focused on the evidence. Tier~1 questions often sound trivial, so models attempt to answer from prior knowledge, skip the tools, and either hallucinate or get marked wrong for lack of a boxed citation. Tier~2 prompts fall into an awkward middle ground: they rarely demand a map, but they do need two textual sources, and our 10-step budget is just enough for the model to read one file before stalling. As a result, we see higher completion rates on the clearly multi-hop Tier~3 set than on apparently “easy” Tier~1/2 questions where the agent underestimates the amount of retrieval required.

\subsection{Experimental Setup}

We evaluate a 60-question slice of MilSCORE drawn from the OPORD scenario, spanning the three difficulty tiers. The evaluation agent is a tool-using chain-of-thought program that can: (i) list available files, (ii) read OPORD PDFs (bounded \texttt{page\_limit}), (iii) query spreadsheets, and (iv) load maps and images. Images are downscaled to a maximum side of 1024~px and injected natively into the prompt. We do not pass any ``input data source'' hints from the CSV - the model must discover sources via tools. All runs use temperature~0, a shared system prompt that asks for a boxed final answer, and a cap of 10 tool/think steps.

\paragraph{Tooling and protocol.} The agent follows a simple ReAct loop: the model proposes a tool name with JSON arguments; the runtime executes the tool and immediately appends a \texttt{ToolMessage} that references the associated \texttt{tool\_call\_id}; only then does the model see the observation and choose its next step. To satisfy vision API requirements, when the model calls \texttt{get\_image\_data} we both (a) return a \texttt{ToolMessage} acknowledging success, and (b) append a new user message that carries the downscaled image as a base64 data URL so the model can actually ``see'' the map. The OPORD reader trims to a bounded page limit per call and returns plain text; the spreadsheet tool returns a short preview (first rows/columns) to keep context small.

\paragraph{Discovery policy.} The model receives no per-question pointers; it is expected to call \texttt{list\_files} first, pick candidate sources by filename, and verify relevance by skimming. We enforce the provider tool-calling contract (every assistant tool call is followed by a matching \texttt{ToolMessage}) and treat any API/protocol/tool errors as incorrect.

We report results for three production VLMs representative of current practice: GPT-4o, Claude Sonnet, and Google's PaliGemma~2.

\subsection{Evaluation Methods}

We employ OpenAI's GPT-5 model to score an item correct if the agent's final boxed answer, $\boxed{\cdot}$, matches the expert reference under a simple normalization (lowercase, punctuation/whitespace stripped) using substring containment. If no box is produced, we fall back to the final message. Any API, protocol, or tool errors are counted as incorrect. Table~\ref{tab:complexity-acc} summarizes the preliminary results on the 60-question slice, reporting for each model the number of correctly answered questions in each difficulty tier together with overall accuracy.

\section{Conclusion}

We introduced MilSCORE, a long-context benchmark for geospatial reasoning and military decision-making grounded in realistic operation order materials. The dataset organizes expert-authored questions into three difficulty tiers and seven spatial-analysis categories, and is paired with a tool-using chain-of-thought evaluation protocol. Our preliminary experiments with several state-of-the-art vision-language models show that, despite strong general capabilities, they still struggle with scenario-level military planning, especially on higher-tier multi-hop questions. We hope MilSCORE will serve as a challenging testbed for developing more reliable, spatially aware long-context models.

\bibliographystyle{abbrv}
\bibliography{references}

\appendix
\section{Appendix - Sample Multi-Hop Trace}
\label{app:long-trace}

To illustrate a longer reasoning chain, we show Question~9 from the Claude 3.5 Haiku run (``Based on the distribution of the permanent population, determine among Apple Valley, Lucerne Valley and Johnson Valley, where is the suitable location for the troops to be stationed.''). The agent iteratively inspects spreadsheets, OPORD annexes, and a terrain map before returning the boxed conclusion.

\begin{scriptsize}
\begin{verbatim}
Processing Q9: Based on the distribution of the permanent population, determine among
Apple Valley, Lucerne Valley and Johnson Vally, where is the suitable location for the
troops to be stationed.
--- Starting RAGNAR Agent for Query:
    Based on the distribution of the permanent population, determine among
    Apple Valley, Lucerne Valley and Johnson Vally, where is the suitable
    location for the troops to be stationed. ---
[Executing Tool]: list_files with args {}
[Executing Tool]: query_spreadsheet with args
    {'filename': '25ID_OPORD_2029-25_location_coordinates.xlsx'}
[Executing Tool]: read_document_text with args
    {'filename': 'Desert_Tortoise_25ID_OPORD.docx'}
[Executing Tool]: read_document_text with args
    {'filename': '25ID_WO1_OPORD_2029-25.docx'}
[Executing Tool]: get_image_data with args
    {'filename': 'terrain_2-1ad_overview.png'}
[System] Injected image as User Message.
[Executing Tool]: query_spreadsheet with args
    {'filename': '25ID_OPORD_2029-25_location_coordinates.xlsx',
     'query': 'Johnson Valley'}
[System] Detected boxed answer. Terminating.
\end{verbatim}
\end{scriptsize}

The resulting response (abridged) is:

\begin{scriptsize}
\begin{verbatim}
... Johnson Valley ... "wide-open fields of fire and almost no urban area,"
indicating the lowest permanent population among the three options.

\boxed{Johnson Valley is the suitable location for troops to be stationed.
It has the lowest permanent population ... while minimizing civilian exposure.}
\end{verbatim}
\end{scriptsize}

\end{document}